\documentclass[letterpaper, 10 pt, conference]{ieeeconf}  

\IEEEoverridecommandlockouts                              

\newcommand\blfootnote[1]{%
    \begin{NoHyper}%
  \begingroup
  \renewcommand\thefootnote{}\footnote{#1}%
  \addtocounter{footnote}{-1}%
  \endgroup
    \end{NoHyper}%
}

\usepackage{graphics} 
\usepackage{epsfig} 
\usepackage{amsmath} 
\DeclareMathOperator*{\argmax}{arg\,max}

\usepackage{amsfonts}
\usepackage{multirow}
\usepackage[utf8]{inputenc}
\usepackage{amssymb}  
\def\BibTeX{{\rm B\kern-.05em{\sc i\kern-.025em b}\kern-.08em
    T\kern-.1667em\lower.7ex\hbox{E}\kern-.125emX}}
\usepackage{subcaption}
\usepackage[font=small,tableposition=top]{caption}
\usepackage[ruled,vlined,linesnumbered]{algorithm2e}
\SetKwInput{KwInput}{Input}

\makeatletter
\let\NAT@parse\undefined
\makeatother
\usepackage{hyperref}  

\title{\LARGE \bf
MILD: Multimodal Interactive Latent Dynamics\\for Learning Human-Robot Interaction}

\author{Vignesh Prasad$^{1,2}$, Dorothea Koert$^{1,5}$, Ruth Stock-Homburg$^2$, Jan Peters$^{1,3,4,5}$, Georgia Chalvatzaki$^{1,4}$
}

\begin{document}

\maketitle
\thispagestyle{empty}
\pagestyle{empty}

\begin{abstract}
Modeling interaction dynamics to generate robot trajectories that enable a robot to adapt and react to a human's actions and intentions is critical for efficient and effective collaborative Human-Robot Interactions (HRI). Learning from Demonstration (LfD) methods from Human-Human Interactions (HHI) have shown promising results, especially when coupled with representation learning techniques. However, such methods for learning HRI either do not scale well to high dimensional data or cannot accurately adapt to changing via-poses of the interacting partner. We propose Multimodal Interactive Latent Dynamics (MILD), a method that couples deep representation learning and probabilistic machine learning to address the problem of two-party physical HRIs. We learn the interaction dynamics from demonstrations, using Hidden Semi-Markov Models (HSMMs) to model the joint distribution of the interacting agents in the latent space of a Variational Autoencoder (VAE). Our experimental evaluations for learning HRI from HHI demonstrations 
show that MILD effectively captures the multimodality in the latent representations of HRI tasks, allowing us to decode the varying dynamics occurring in such tasks. Compared to related work, MILD generates more accurate trajectories for the controlled agent (robot) when conditioned on the observed agent's (human) trajectory. Notably, MILD can learn directly from camera-based pose estimations to generate trajectories, which we then map to a humanoid robot without the need for any additional training. \\Supplementary Material: \url{https://bit.ly/MILD-HRI}

\end{abstract}
\vspace{-1.5em}

\blfootnote{
\hspace{-1em}$^{1}$ Department of Computer Science, TU Darmstadt, Germany.\\
$^{2}$ Chair for Marketing and Human Resource Management, Department of Law and Economics, TU Darmstadt, Germany.\\
$^{3}$ German Research Center for AI (DFKI), Research Department: Systems AI for Robot Learning.\\
$^{4}$ Hessian.AI\\
$^{5}$ Centre for Cognitive Science, TU Darmstadt, Germany.\\
This work was supported by the German Research Foundation (DFG) Project "Social Robots at the Customer Interface" (Grant No.: STO 477/14-1), the DFG Emmy Noether Programme (CH 2676/1-1), the German Federal Ministry of Education and Research (BMBF) Project "IKIDA" (Grant no.: 01IS20045), the RoboTrust project of the Centre Responsible Digitality (ZEVEDI) Hessen, Germany, the Funding Association for Market-Oriented Management, Marketing, and Human Resource Management (F\"orderverein f\"ur Marktorientierte Unternehmensf\"uhrung, Marketing und Personal management e.V.), and the Leap in Time Foundation (Leap in Time Stiftung). The authors thank the NHR Centers NHR4CES at TU Darmstadt for the access to the Lichtenberg high-performance computer (Project No. 1694) for running the experiments in this work.}
\section{Introduction}


Observing human actions and interacting synchronously is an essential characteristic of a social robot in HRI scenarios \cite{ajoudani2018progress}. Key components for learning coordinated HRI policies are having a good spatio-temporal representation and jointly modeling the interaction dynamics of the agents. In this regard, the paradigm of LfD shows promising results~\cite{amor2014interaction,campbell2017bayesian,calinon2009learning}, especially when using only a handful of trajectories. Such LfD approaches learn joint distributions over human and robot trajectories that can be conditioned on the observed actions of the human, although, they scale poorly with higher dimensions. In such cases, deep LfD approaches perform well for learning latent-space dynamics with high-dimensional data~\cite{karl2016deep,chen2016dynamic,chaveroche2018variational,dermy2018prediction,nagano2019hvgh} 
but they usually are not scalable to different interactive scenarios, as they usually do not model the inherent multimodality and uncertainty of HRI tasks.

\begin{figure}[t!]
    \centering
    \includegraphics[width=\linewidth]{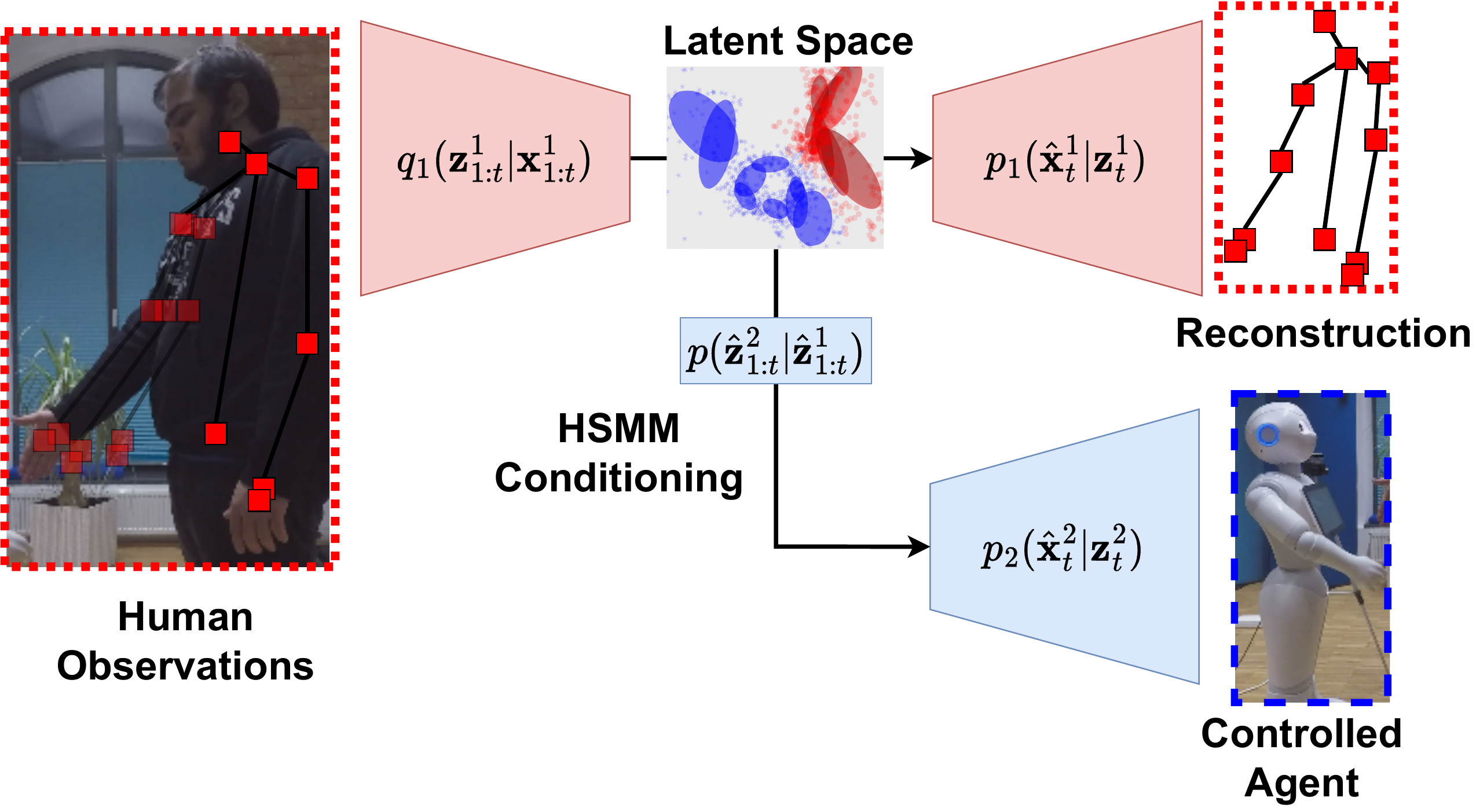}
    \caption{Overview of our approach, "MILD". We train VAEs to reconstruct the observations of the interactions agents $(\boldsymbol{x}^1_{1:t},\boldsymbol{x}^2_{1:t})$ with an HSMM prior to learn a joint distribution over the latent space trajectories $(\boldsymbol{z}^1_{1:t},\boldsymbol{z}^2_{1:t})$ of the interacting agents. During test time, the observed agent's latent trajectory conditions the HSMM to infer the controlled agent's latent trajectory $p(\boldsymbol{z}^2_t|\boldsymbol{z}^1_{1:t})$ which is decoded to generate the agent's real-world trajectory $\boldsymbol{\hat{x}}^2_t$.}
    \label{fig:vaehsmm-test}
    \vspace{-2em}
\end{figure}

To tackle these challenges when learning HRI policies, we introduce MILD, a method that effectively couples benefits from deep LfD methods with probabilistic machine learning. MILD uses HSMMs as a temporally coherent latent space prior of a VAE, which we learn from HHI data. Specifically, we model the prior as a joint distribution over the trajectories of both interacting agents, making full use of the power of HSMMs for learning both trajectory and interaction dynamics. Our approach successfully captures the multimodality of the latent interactive trajectories thanks to the modularity of HSMMs, enabling better modeling of dynamics as compared to using an uninformed, stationary prior~\cite{butepage2020imitating}. 
During testing, we can generate latent trajectories by conditioning the HSMM on the human observations using Gaussian Mixture Regression~\cite{calinon2009learning}, and decode them to obtain the robot's control trajectories (Fig. \ref{fig:vaehsmm-test}). 

Our experimental evaluations on different test scenarios show the efficacy of MILD in capturing coherent latent dynamics, both in predicting HHI and in generating controls for HRI on different robots, 
compared to the state-of-the-art method that implicitly learns shared representations~\cite{butepage2020imitating}. MILD learns to generate effective robot trajectories for HRI, not only through robot kinesthetic teaching, but also by learning from HHI, both from idealistic data (Motion Capture) and noisy RGB-D skeleton tracking, by directly transferring the generated trajectories to a humanoid robot~\cite{fritsche2015first,prasad2021learning}, without requiring additional demonstrations or fine-tuning.


\section{Related Work}
\label{sec:related}
Early approaches for learning HRI modeled them as a joint distribution with a Gaussian Mixture Model (GMM) learned over demonstrated trajectories of a human and a robot in a collaborative task~\cite{calinon2009learning}. The correlations between the human and the robot degrees of freedom (DoFs) can then be leveraged to generate the robot's trajectory given observations of the human. This method was further extended with HSMMs with explicit duration constraints for learning both proactive and reactive controllers~\cite{rozo2016learning}. Along similar lines of leveraging Gaussian approximations for LfD, Movement primitives~\cite{paraschos2013probabilistic,schaal2006dynamic}, which learn a distribution over underlying weight vectors obtained via linear regression, were extended for HRI by similarly learning a joint distribution over the weights of both interacting agents~\cite{amor2014interaction,maeda2014learning}. 
The versatility of interaction primitives can additionally be noted by their ability to be adapted for different intention predictions~\cite{koert2019learning}, speeds~\cite{maeda2017phase}, or for learning multiple interactive tasks seamlessly by either using a GMM as the underlying distribution~\cite{ewerton2015learning} or in an incremental manner~\cite{maeda2017active,koert2018online}.

Deep LfD techniques have grown in popularity for learning latent trajectory dynamics from demonstrations wherein an autoencoding approach, like VAEs, is used to encode latent trajectories over which a latent dynamics model is trained. In their simplest form, the latent dynamics can be modeled either with linear Gaussian models~\cite{karl2016deep} or Kalman filter~\cite{becker2019recurrent}. Other approaches learn stable dynamical systems, like Dynamic Movement Primitives~\cite{schaal2006dynamic} over VAE latent spaces~\cite{bitzer2009latent,chen2015efficient,chen2016dynamic, chaveroche2018variational}. Instead of learning a feedforward dynamics model, Dermy et al.~\cite{dermy2018prediction} model the entire trajectory's dynamics at once using Probabilistic Movement Primitives~\cite{paraschos2013probabilistic} achieving better results than~\cite{chaveroche2018variational}. When large datasets are available, Recurrent Networks are powerful tools in approximating latent dynamics~\cite{chung2015recurrent,han2021disentangled}, especially in the case of learning common dynamics models in HRIs~\cite{butepage2020imitating}. A major advantage of most of the aforementioned LfD approaches, other than their sample efficiency in terms of demonstrations, is that they can be explicitly conditioned at desired time steps, unlike neural network-based approaches.

Most deep LfD approaches fit complete trajectories, curating neither the multimodality in HRI tasks nor the subsequent dynamics between different modes of interaction. Instead of fitting a single distribution over demonstrated trajectories, HSMMs break down such complex trajectories into multiple modes and learn the sequencing between hidden states, as shown in~\cite{nagano2019hvgh}, where HSMMs were used as latent priors for a VAE. However, \cite{nagano2019hvgh} does not look at the interdependence between dimensions, but models each dimension individually, which is not favorable when learning interaction dynamics. Such issues can be circumvented by using a diagonal cross-covariance structure (as in~\cite{becker2019recurrent}), but this would only learn dependencies between individual dimensions. Contrarily, we learn full covariance matrices in our HSMM models to better facilitate the learning of interaction dynamics.

\section{Preliminaries}
\label{sec:preliminaries}
In this section, we briefly introduce preliminary concepts, namely, VAEs (Sec.~\ref{ssec:ae-vae}) and HSMMs (Sec.~\ref{ssec:hsmm}), that we deem useful for discussing our proposed method.
\subsection{Variational Autoencoders}
\label{ssec:ae-vae}

Variational Autoencoders (VAEs)~\cite{kingma2013auto,rezende2014stochastic} are a style of neural network architectures that learn the identity function in an unsupervised, probabilistic way. An encoder encodes the inputs onto a latent space, denoted by "$\boldsymbol{z}$", of the input "$\boldsymbol{x}$" at the bottleneck that a decoder uses to reconstruct the original input. A prior distribution is enforced over the latent space, usually given by a normal distribution $p(\boldsymbol{z})=\mathcal{N}(\boldsymbol{z}; \boldsymbol{0, I})$. The goal is to estimate the true posterior $p(\boldsymbol{z|x})$, using a neural network $q(\boldsymbol{z|x})$ and is trained by minimizing the Kullback-Leibler (KL) divergence between them.

\begin{equation}
    D_{KL}(q(\boldsymbol{z|x})||p(\boldsymbol{z|x})) = \mathbb{E}_q[\log\frac{ q(\boldsymbol{z|x})}{p(\boldsymbol{x,z})}] + \log p(\boldsymbol{x})
\end{equation}
This can be re-written as 
\begin{equation}
    \label{eq:evidence_log}
    \log p(\boldsymbol{x}) = D_{KL}(q(\boldsymbol{z|x})||p(\boldsymbol{z|x})) + \mathbb{E}_q[\log\frac{p(\boldsymbol{x,z})}{q(\boldsymbol{z|x})}]
\end{equation}

The KL divergence is always non-negative, therefore the second term in \eqref{eq:evidence_log} acts as a lower bound. Maximizing it would effectively maximize the log-likelihood of the data distribution or evidence, and is hence called the Evidence Lower Bound (ELBO), which can be written as
\begin{equation}
    \label{eq:elbo}
    \mathbb{E}_q[\log\frac{p(\boldsymbol{x,z})}{q(\boldsymbol{z|x})}] = \mathbb{E}_q[\log p(\boldsymbol{x|z})] + D_{KL}(q(\boldsymbol{z|x})||p(\boldsymbol{z}))
\end{equation}
The first term corresponds to the reconstruction of the input via samples decoded from the posterior distribution. The second term is the KL divergence between the prior and the approximate posterior, which acts as a regularization term for the posterior. Further information about variational inference can be found in~\cite{kingma2013auto,rezende2014stochastic}.

\begin{figure*}[ht!]
\vspace{1.5em}
    \centering
        \includegraphics[width=0.75\linewidth]{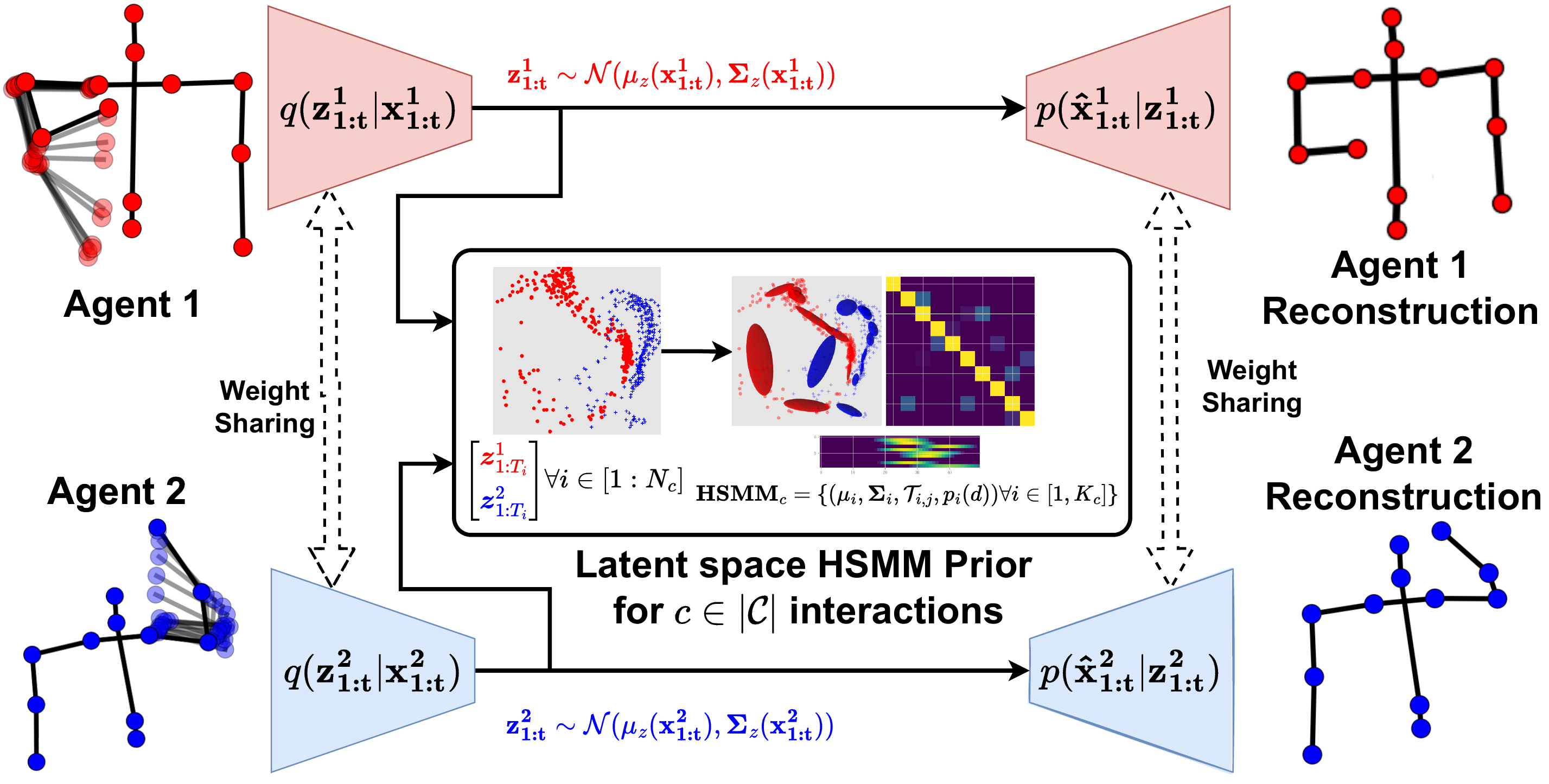}
    \caption{Pipeline for training MILD: We model the interaction dynamics between two agents in the latent space of a VAE using a temporally coherent prior in the form of an HSMM to model a joint distribution over the latent trajectories. To match the predictive abilities of the HSMM prior, we predict full posterior covariance matrices, rather than forcing a diagonalized covariance like traditional VAEs. The use of an HSMM prior, as opposed to standard Gaussians or other LfD approaches, enforces a modularized latent space using a GMM ($\boldsymbol{\mu}_i, \boldsymbol{\Sigma}_i$) which subsequently learns the transitions between clusters $\mathcal{T}_{i,j}$ as well as duration statistics $p_i(d)$ to improve the transition prediction.
    \label{fig:vaehsmm-train}
    }
    \vspace{-2em}
\end{figure*}

\subsection{Hidden Semi-Markov Models}
\label{ssec:hsmm}
HSMMs are a special class of Hidden Markov Models (HMMs), where the Markov property is relaxed, i.e., the current state depends on not just the previous state, but also the duration for which the system remains in a state. In an HMM, a sequence of observations $\boldsymbol{z}_{1:T}$ is modeled as a sequence of $K$ hidden latent states that emit the observations with some probability. Specifically, an HMM can be described by its initial state distribution $\pi_i$ over the states $i\in\{1, 2 \dots K\}$, the state transition probabilities $\mathcal{T}_{i,j}$ denoting the probability of transitioning from state $i$ to state $j$. In our case, each state is characterized by a Normal distribution with mean $\boldsymbol{\mu}_i$ and covariance $\boldsymbol{\Sigma}_i$, which characterize the emission probabilities of the observations $\mathcal{N}(\boldsymbol{z}_t;\boldsymbol{\mu}_i, \boldsymbol{\Sigma}_i)$. This, in essence, is similar to learning a GMM over the observations and learning the temporal sequencing between the Gaussian components, by learning the forward variable of the HMM $h_i(\boldsymbol{z}_t)$
\begin{equation}
\label{eq:hmm-h}
    h_i(\boldsymbol{z}_t) = \alpha_i(\boldsymbol{z}_t)/\sum\nolimits_{k=1}^K\alpha_k(\boldsymbol{z}_t)
\end{equation}
where
\begin{equation}
\label{eq:hmm-alpha}
\alpha_i(\boldsymbol{z}_t) = \mathcal{N}(\boldsymbol{z}_t;\boldsymbol{\mu}_i, \boldsymbol{\Sigma}_i)\sum_{k=1}^K\alpha_k(\boldsymbol{z}_{t-1})\mathcal{T}_{k,i}
\end{equation}
and $\alpha_i(\boldsymbol{z}_0) = \pi_i$. In the case of HSMMs, a distribution $p_i(d)$ is additionally fitted over the number of steps $d\in\{1,2,\dots D\}$ that the model stays in a given state
\begin{equation}
\label{eq:hsmm-alpha}
    \alpha_i(\boldsymbol{z}_t) = \mathcal{N}(\boldsymbol{z}_t;\boldsymbol{\mu}_i, \boldsymbol{\Sigma}_i)\sum_{k=1}^K\sum_{d=1}^D\alpha_k(\boldsymbol{z}_{t-1})\mathcal{T}_{k,i}p_i(d)
\end{equation}

For a more in-depth explanation of training GMMs, HMMs and HSMMs in the context of robot learning, we refer the reader to~\cite{calinon2016tutorial,pignat2017learning}.
To encode the joint distribution between the interacting agents, an observation space is constructed by concatenating the DoFs of both agents~\cite{calinon2009learning,evrard2009teaching}. This allows the distribution to be decomposed as
\begin{equation}
\label{eq:gmr}
    \boldsymbol{\mu}_i = \begin{bmatrix}
\boldsymbol{\mu}^1_i\\
\boldsymbol{\mu}^2_i
\end{bmatrix}; \boldsymbol{\Sigma}_i = \begin{bmatrix}
\boldsymbol{\Sigma}^{11}_i & \boldsymbol{\Sigma}^{12}_i\\
\boldsymbol{\Sigma}^{21}_i & \boldsymbol{\Sigma}^{22}_i
\end{bmatrix}
\end{equation}

Once the distributions are learned, given observations $\boldsymbol{z}^1_{1:t}$ of Agent 1, the trajectory for Agent 2 $\boldsymbol{z}^2_{1:t}$ is inferred as
\begin{equation}
    \label{eq:gmr-conditioning}
    \boldsymbol{z}^2_{1:t} = \sum_{k=1}^K h_k(\boldsymbol{z}^1_{1:t}) (\boldsymbol{\mu}^2_k + \boldsymbol{\Sigma}^{21}_k(\boldsymbol{\Sigma}^{11}_k)^{-1}(\boldsymbol{\mu}^1_i - \boldsymbol{z}^1_{1:t})).
\end{equation}

This allows the HSMM to adapt the controlled agent's trajectory to the observed agent thereby capturing the interaction dynamics and resulting in more accurate interaction.

\section{Multimodal Interactive Latent Dynamics}
\label{sec:approach}
In this section, we introduce our approach, MILD (Multimodal Interactive Latent Dynamics), which uses the representation learning abilities of VAEs coupled with the abilities of HSMMs for interaction modeling by incorporating the HSMMs as the prior in the VAE. Our method extends the idea proposed in~\cite{nagano2019hvgh} who showed promising results for trajectory segmentation using an HSMM prior with a VAE. However, in \cite{nagano2019hvgh}, the correlation between different dimensions is not modeled, which is a key factor when jointly modeling the interaction dynamics in HRI scenarios. Therefore, rather than learning a single model per dimension, we model full covariance matrices in the HSMM, allowing us to predict the inter-dependency between human and robot DOFs. We describe the approach of training a VAE with an HSMM prior in Sec.~\ref{ssec:vae-hsmm}, which is visualized in Fig.~\ref{fig:vaehsmm-train}, followed by how the robot actions can be interactively generated by conditioning the HSMM in Sec.~\ref{ssec:hsmm_conditioning}.

\subsection{Learning VAEs with HSMMs priors for HRI}
\label{ssec:vae-hsmm}
Typically in VAEs, the prior $p(\boldsymbol{z})$ is modeled as a stationary distribution that doesn't change with time. When it comes to learning trajectories, this can be a very hard constraint. We argue that having meaningful transition priors can help learn temporally coherent latent spaces~\cite{chen2016dynamic}. Modeling trajectory priors could potentially be achieved by autoencoding sub-sequences rather than individual time-steps like in~\cite{butepage2020imitating}, but it still requires learning a temporal dynamics model to represent the progression of the trajectory. To this end, we explore the use of HSMMs that not only learn latent-space dynamics but are also able to break down the interactions into multiple phases and learn their sequencing, leading to a more modular structure, and can also learn inter-dependencies between actions. We defer this direction of jointly learning action segmentation and the underlying modular skills to future work, and we mainly focus on learning latent-space HSMMs for HRI to capture the task-wise multimodality. 

In order to capture the interactive behavior, we learn a joint distribution over the trajectories of both interacting agents. Given a latent encoding of the trajectories of both interacting agents $\boldsymbol{z}^1_{1:T}$ and $\boldsymbol{z}^2_{1:T}$, the prior probabilities of the encoded observations at each time-step $\boldsymbol{z}_t$ of an input $\boldsymbol{x}_t$ are calculated using the current most likely component of the HSMM $\hat{i}$ (without any observation, by dropping the emission probability from~\eqref{eq:hmm-h} and~\eqref{eq:hsmm-alpha}) whose parameters are used as the prior in the ELBO.
\begin{equation}
\label{eq:vaehsmm-prior}
\begin{split}
    \hat{i} &= \argmax_i h_i(\cdot)\\
    p(\boldsymbol{z}^s_t) &= \mathcal{N}(\boldsymbol{z}^s_t;\boldsymbol{\mu}^s_{\hat{i}}, \boldsymbol{\Sigma}^{ss}_{\hat{i}}) ; s\in\{1,2\}
\end{split}
\end{equation}
where $s$ denotes each agent's index.
The HSMM parameters of all the interactions are fixed during a training epoch. After the training of the VAE for an epoch, the new HSMM parameters are estimated using the learned trajectory encodings. 

VAEs model the posterior distribution with a simple diagonalized covariance matrix, whereas the HSMM distributions are modeled with a full covariance matrix that captures the relations between the different DoFs. Therefore, we predict the full covariance matrix as well, to better represent the correlation between dimensions of the demonstrated trajectories. However, this is not a straightforward task, as we need to ensure that the predicted matrix is symmetric and positive definite. While this can be approximated by finding the closest symmetric positive definite matrix within its Frobenius norm~\cite{higham1988computing}, it involves computing the eigendecomposition of the matrix and reconstructing it back for each data point. This makes it computationally expensive, especially when it involves backpropagation through the networks for a large number of samples. Therefore, we follow the approach presented in~\cite{dorta2018structured} by predicting the lower triangular matrix $\boldsymbol{L}$ corresponding to the Cholesky decomposition of the covariance matrix $\boldsymbol{\Sigma} = \boldsymbol{L}\boldsymbol{L}^T$. The covariance matrix can be made positive definite by enforcing the diagonal elements of $\boldsymbol{L}$ to be positive. Our posterior can now be written as $q(\boldsymbol{z}_t|\boldsymbol{x}_t) \sim \mathcal{N}(\boldsymbol{\mu}(\boldsymbol{x}_t), \boldsymbol{\Sigma}(\boldsymbol{x}_t))$, where $\boldsymbol{\mu}(\boldsymbol{x}_t)$ and $\boldsymbol{\Sigma}(\boldsymbol{x}_t) = \boldsymbol{L}(\boldsymbol{x}_t)\boldsymbol{L}^T(\boldsymbol{x}_t)$ are the outputs of a neural network. This 
allows us to model a full covariance matrix that can match the form of the prior. 

Alg.~\ref{alg:vae-hsmm} provides an overview of the training of MILD. Given a set of labelled trajectories $\boldsymbol{X}$ containing demonstrations of two agents $\boldsymbol{X}^1_{1:T}$ and $\boldsymbol{X}^2_{1:T}$ for $\mathcal{C}$ interactions, in each epoch, the VAE is trained by optimizing~\eqref{eq:elbo} using the prior according to~\eqref{eq:vaehsmm-prior}. Subsequently, an HSMM corresponding to each of the $c \in |\mathcal{C}|$ interactions is trained with the demonstrations of that interaction of both agents jointly using Expectation Maximization as explained in~\cite{pignat2017learning}.

\begin{algorithm}[h!]
\small
\vspace{0.1em}
\SetAlgoLined
\KwData{A set of trajectories with action labels $\boldsymbol{X} =\{\boldsymbol{X}^1_{1:T}, \boldsymbol{X}^2_{1:T}, c\}$ for $|\mathcal{C}|$ actions}
 \KwResult{VAE weights and $|\mathcal{C}|$ HSMM parameters}
 Initialize VAE weights randomly\; 
 Initialize $\boldsymbol{\mu}^c_i \gets \boldsymbol{0}, \boldsymbol{\Sigma}^c_i \gets \boldsymbol{I} \ \forall c \in [1,|\mathcal{C}|] \ \forall i \in [1,K_c]$ \
 \While{not converged}{
 \For {$\boldsymbol{x}^1_{1:T}, \boldsymbol{x}^2_{1:T}, c \in \boldsymbol{X}$}{
    for $s\in\{1,2\} $ Maximize $\mathbb{E}_q[\log p_s(\boldsymbol{x}^s_t|\boldsymbol{z}^s_t)] + D_{KL}(q_s(\boldsymbol{z}^s_t|\boldsymbol{x}^s_t)||p_c(\boldsymbol{z}^s_t))$\;\label{alg:line:elbo}
    where $p_c(\boldsymbol{z}^s_t)$ is calculated using Eq.~\ref{eq:vaehsmm-prior}
 }
  \For {$c \in [1,|\mathcal{C}|]$}{
  $\boldsymbol{X}^c \gets$ set of demonstrations of Interaction $c$\;
  $\boldsymbol{Z}^c \gets \emptyset$ \;
  \For {$\boldsymbol{x}^1_{1:T}, \boldsymbol{x}^2_{1:T}, c \in \boldsymbol{X}^c$}{
  $\boldsymbol{z}^1_{1:T}\sim q_1(\cdot|\boldsymbol{x}^1_{1:T}) ; \boldsymbol{z}^2_{1:T}\sim q_2(\cdot|\boldsymbol{x}^2_{1:T})$
  $\boldsymbol{Z}^c \gets \boldsymbol{Z}^c \cup \begin{bmatrix}
\boldsymbol{z}^1_{1:T}\\
\boldsymbol{z}^2_{1:T}
\end{bmatrix}$ \;
  
  }
  Train the $c^{th}$ HSMM with $\boldsymbol{Z}^c$
  }
  
 }
 \caption{Training MILD}
 \label{alg:vae-hsmm}
\end{algorithm}

\subsection{Conditional Trajectory Generation}
\label{ssec:hsmm_conditioning}
During testing, given $t$ observations of the human agent $\boldsymbol{x}^1_{1:t}$, we first encode the observations $\boldsymbol{z}^1_{1:t}\sim q_1(\cdot|\boldsymbol{x}^1_{1:t})$ and then condition the learned joint distribution of the given action using the latent encodings of the observed trajectory. The conditioned HSMM model is then used to generate the latent trajectory of the second agent $\hat{\boldsymbol{z}}^2_{1:t}$ using~\eqref{eq:gmr-conditioning}, which is then decoded to obtain the actions of the second agent $\hat{\boldsymbol{x}}^2_{1:t}\sim p_2(\cdot|\boldsymbol{z}^2_{1:t})$. This process is shown in Alg.~\ref{alg:vae-hsmm-test}. We currently do not perform action recognition to select which HSMM to condition, which we defer to future work.

\begin{algorithm}[h!]
\small
\SetAlgoLined
\KwData{An observation of the human agent $\boldsymbol{x}^1_{1:t}$, Trained MILD Model}
\KwResult{Conditioned Trajectory for the second agent $\hat{\boldsymbol{x}}^2_{1:t}$}
    
    Encode the observed trajectory $\boldsymbol{z}^1_{1:t}\sim q_1(\cdot|\boldsymbol{x}^1_{1:t})$\\
    Condition the HSMM model to get the latent trajectory for the second agent using~\eqref{eq:gmr-conditioning}    $\hat{\boldsymbol{z}}^2_{1:t} = \sum_{k=1}^K h_i(\boldsymbol{z}^1_{1:t}) (\boldsymbol{\mu}^2_i + \boldsymbol{\Sigma}^{21}_i(\boldsymbol{\Sigma}^{11}_i)^{-1}(\boldsymbol{\mu}^1_i - \boldsymbol{z}^1_{1:t}))$\\
    Decode the conditioned trajectory $\hat{\boldsymbol{x}}^2_{1:t}\sim p_2(\cdot|\hat{\boldsymbol{z}}^2_{1:t})$
 \caption{Conditioning on Human Observations}
 \label{alg:vae-hsmm-test}
\end{algorithm}

\begin{figure*}[h!]
\vspace{1em}
\centering
    \begin{subfigure}[b]{0.17\textwidth}
        \centering
        \includegraphics[width=\linewidth]{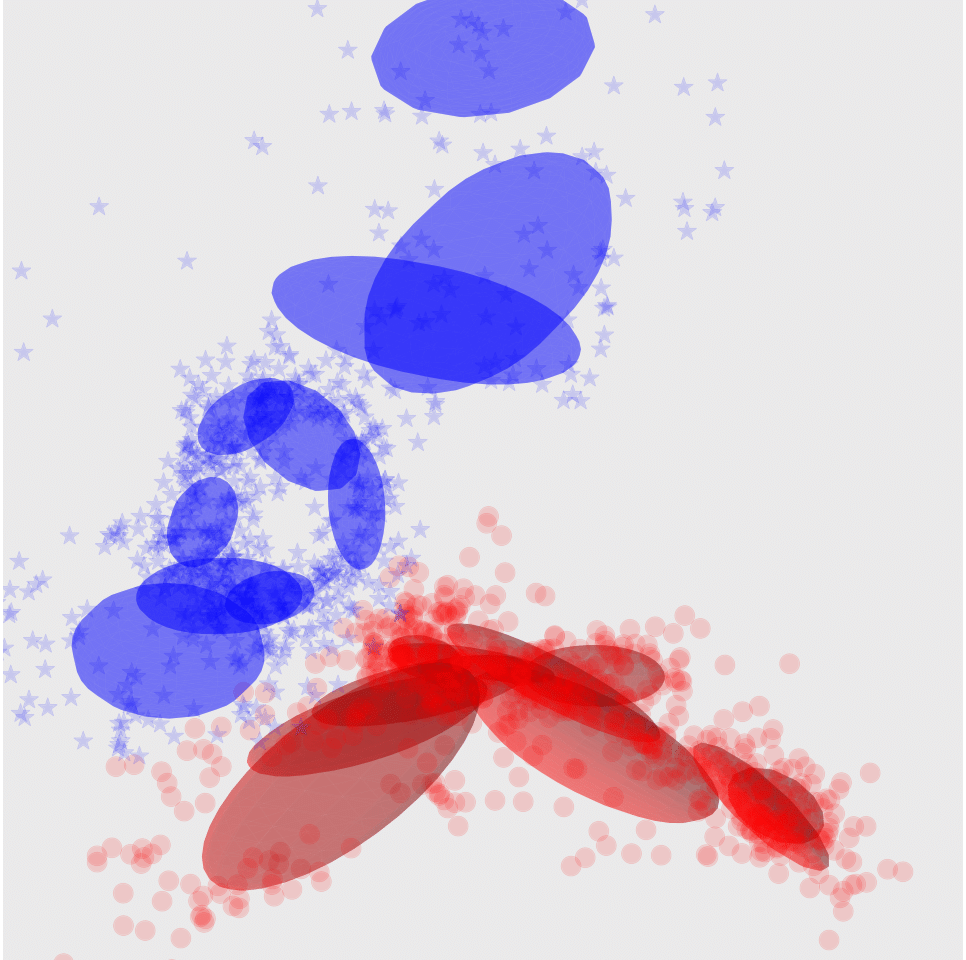}
        \caption{Handshaking}
        \label{fig:handshaking-latent}
    \end{subfigure}
    \begin{subfigure}[b]{0.17\textwidth}
        \centering
        \includegraphics[width=\linewidth]{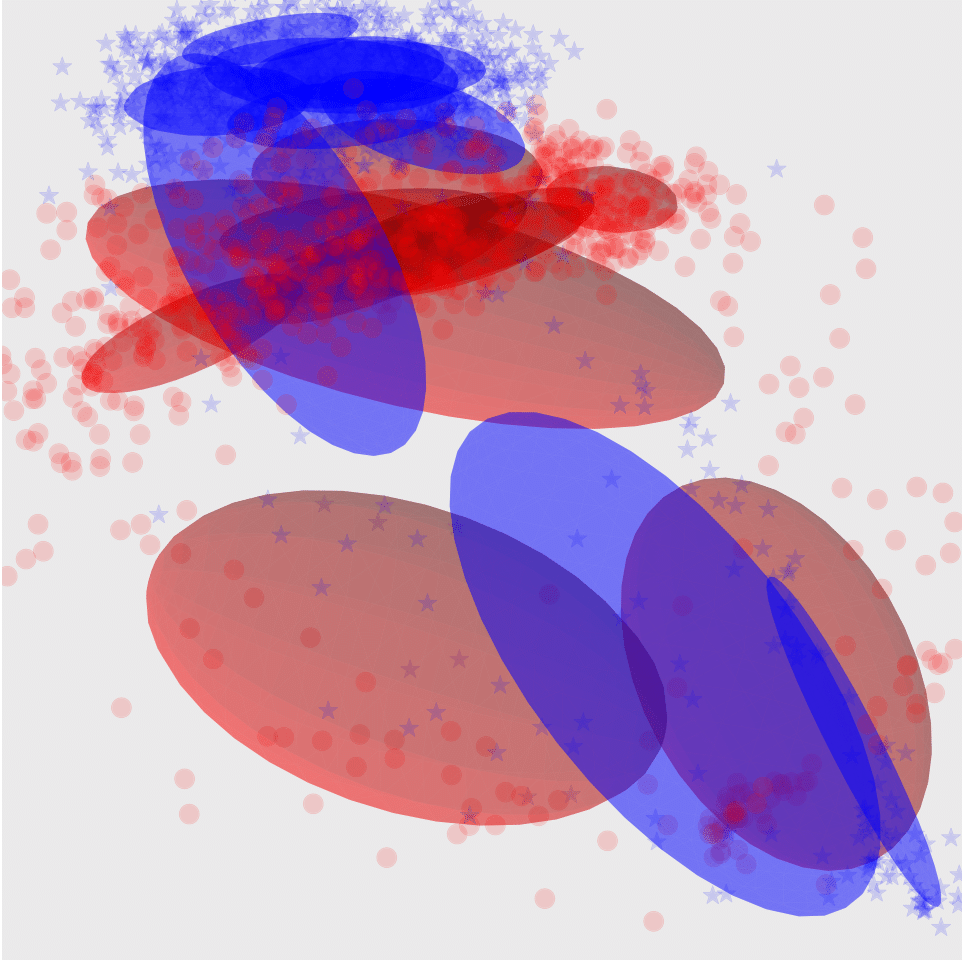}
        \caption{Hand Waving}
        \label{fig:handwaving-latent}
    \end{subfigure}
    \begin{subfigure}[b]{0.17\textwidth}
        \centering
        \includegraphics[width=\linewidth]{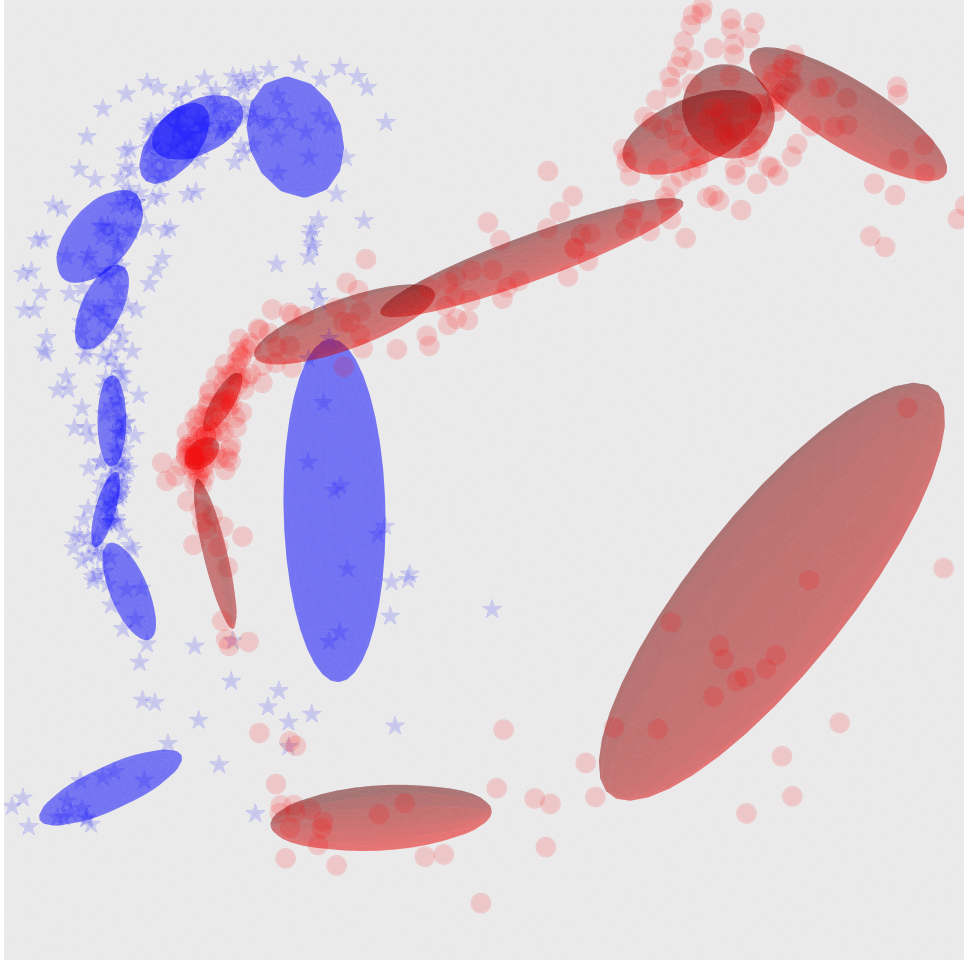}
        \caption{Rocket Fistbump}
        \label{fig:handwaving-latent}
    \end{subfigure}
    \begin{subfigure}[b]{0.17\textwidth}
        \centering
        \includegraphics[width=\linewidth]{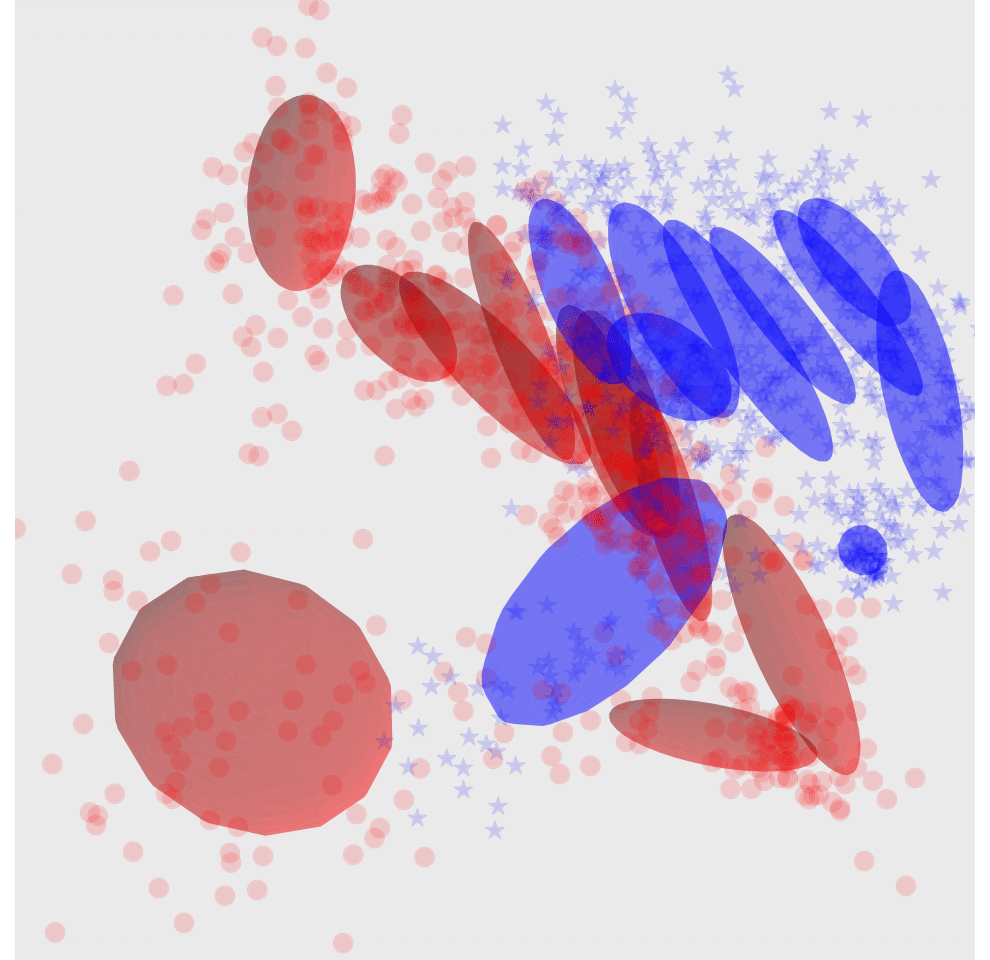}
        \caption{Parachute Fistbump}
        \label{fig:handshaking-pred}
    \end{subfigure}
    \caption{3D Latent spaces learned by MILD on the different interactions in~\cite{butepage2020imitating}. (blue stars - Agent 1, red circles - Agent 2)}
        \label{fig:latents}
        \vspace{-2em}
\end{figure*}

\section{Experiments and Results}
\label{sec:experiments}
In this section, we first explain our setup's implementation (Sec.~\ref{ssec:setup}) and the datasets used (Sec.~\ref{ssec:datasets}), following which we present some qualitative results of the learned latent spaces (Sec.~\ref{ssec:learnt_repr}). Finally, we present the results of predicting the controlled agent's trajectories after conditioning on the observed agent in Sec.~\ref{ssec:motion_pred}.

\subsection{Experimental Setup}
\label{ssec:setup}
The networks are implemented using PyTorch 
with a similar structure as in~\cite{butepage2020imitating} 
Both the encoder and the decoder consist of 2 feedforward hidden layers with dimensions [250, 150] with Leaky ReLU activations 
and 5-dimensional latent space. 
The encoder has two output networks, one for the mean $\boldsymbol{\mu}_{\boldsymbol{z}}$ and one for the lower triangular matrix $\boldsymbol{L}_{\boldsymbol{z}}$ corresponding to the Cholesky decomposition of the covariance matrix $\boldsymbol{\Sigma}_{\boldsymbol{z}} = \boldsymbol{L}_{\boldsymbol{z}}\boldsymbol{L}_{\boldsymbol{z}}^T$. The diagonal elements are forced to be positive as $l^{ii}_{\boldsymbol{z}} = 2|l^{ii}_{\boldsymbol{z}}|$.
To stabilize the VAE training and prevent issues from over regularization, we multiply the KL loss with a scale factor of $10^{-3}$. We additionally decode 10 samples from the posterior to train the networks rather than a single sample, which we found helps improve the training. 
The networks are trained with a learning rate of $10^{-4}$ using the AdamW optimizer. 
The HSMMs are implemented using PbDLib\footnote{\url{https://gitlab.idiap.ch/rli/pbdlib-python}}~\cite{pignat2017learning} 
with 10 GMM components (chosen empirically) for each HSMM. The HSMMs are initialized by splitting the data temporally into equal components.

\subsection{Dataset}
\label{ssec:datasets}
\subsubsection{B\"utepage et al.~\cite{butepage2020imitating}}\footnote{\url{https://github.com/jbutepage/human_robot_interaction_data}} capture their HHI data using a Rokoko Smart suit 
for the human skeleton data and use kinesthetic teaching on an ABB YuMi-IRB 14000 robot for collecting Human-Robot (HRI) data. It consists of demonstrations of 4 actions: Hand Waving, Handshaking, and two different kinds of fist bumps. The first is called Rocket Fistbump, which involves bumping the fists at a low level, followed by raising the fists upwards while in contact with each other. The second is called Parachute Fistbump in which partners bump their fists at a high level near the head and bring it down with simultaneously oscillating the hands sideways, while in contact with each other. A more detailed explanation of the interactions can be found in~\cite{butepage2020imitating}. 

In total, there are 149 trajectories for training and 32 for testing from the HHI setting and 32 training and 9 testing for the HRI setting, as in~\cite{butepage2020imitating}. The trajectories are pre-processed to use the 3D coordinates of 4 right arm joints (as seen in Fig.~\ref{fig:pred-butepage} and~\ref{fig:yumi-handshake}), with the origin at the shoulder. Like in~\cite{butepage2020imitating}, we concatenate 40 observations corresponding to a time window of 1 second, leading to an input size of 480 dimensions (40x4x3). For the Robot trajectories, the data is similarly sampled, leading to an input size of 280 dimensions (40x7) for the 7 joint angles of the robot's right arm. The hidden layers of the Robot VAE are initialized using the weights from the Human VAE to accelerate the learning given the lower number of robot trajectory samples.

\subsubsection{Nuitrack Skeleton Interaction Dataset} (NuiSI) is a dataset that we collected ourselves, which consists of 6 interactions, namely Handshaking, Hand waving, High Fives, Fist bumps, Clap fist (Handclap followed by fist bump), and Rocket Fist bump (like in~\cite{butepage2020imitating}). The data is recorded using an Intel Realsense D435 which provides RGB-D images using Nuitrack\footnote{\url{https://nuitrack.com/}}
for tracking the upper body skeleton joints in each frame. The data is first cleaned to remove any missing trajectories, and the trajectories of both agents are manually aligned and down-sampled to match the trajectory of shorter length. For training, the data is processed similarly as mentioned above with a window size of 5 time steps given the lower frame rate and irregularities during skeleton tracking. We use the pre-trained network trained on the aforementioned skeleton data as well, to help accelerate the learning. We additionally evaluate MILD on the humanoid robot Pepper~\cite{pandey2018mass} after training it on the Nuitrack data by making use of the similar DoFs between a human and a humanoid~\cite{fritsche2015first,prasad2021learning} to extract the joint angles for the robot. We use a moving average filter to reduce the jerk of the generated robot trajectories.





\subsection{Learned Representations}
\label{ssec:learnt_repr}
\begin{figure*}[h!]
\vspace{1.5em}
\centering
    \begin{subfigure}[b]{0.24\textwidth}
        \centering
        \includegraphics[width=\linewidth]{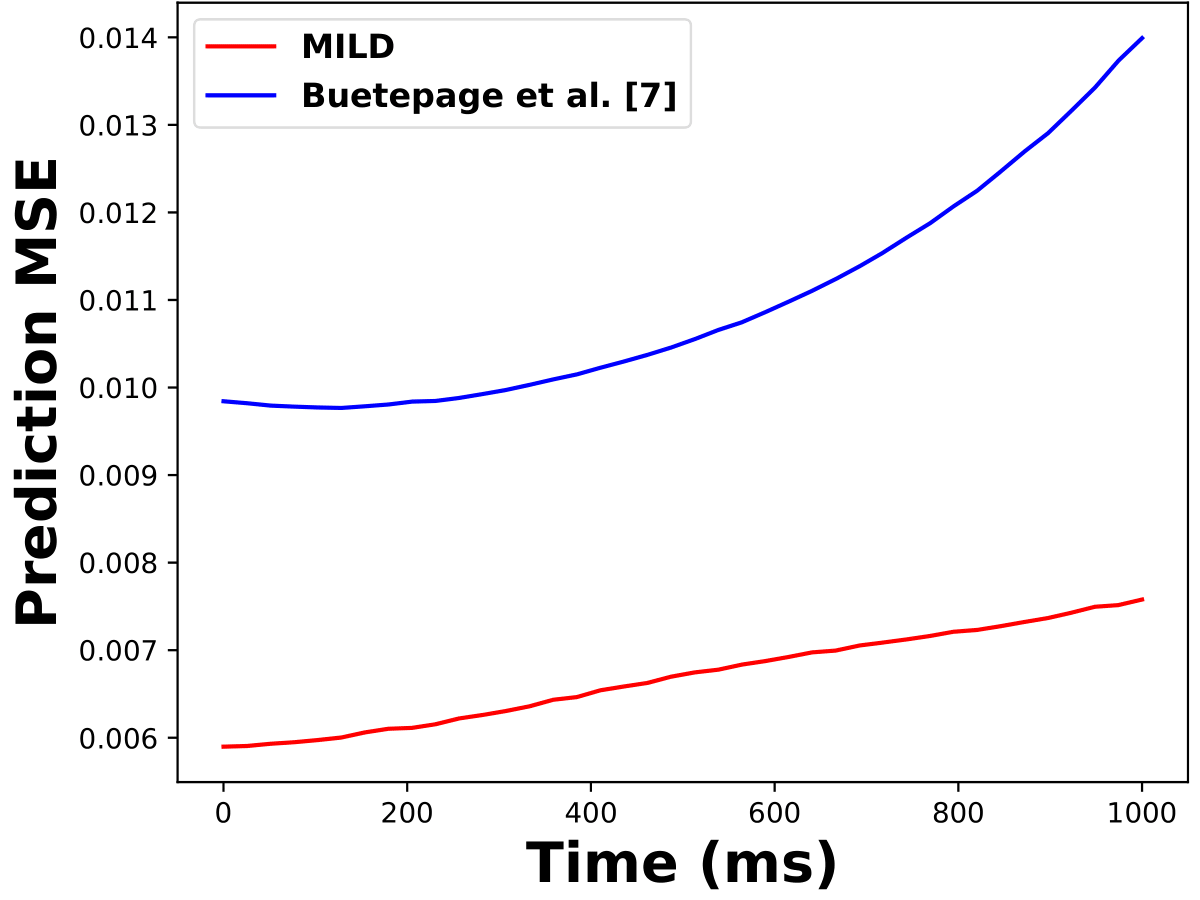}
        \caption{Hand Waving}
        \label{fig:handwaving-pred}
    \end{subfigure}
    \begin{subfigure}[b]{0.24\textwidth}
        \centering
        \includegraphics[width=\linewidth]{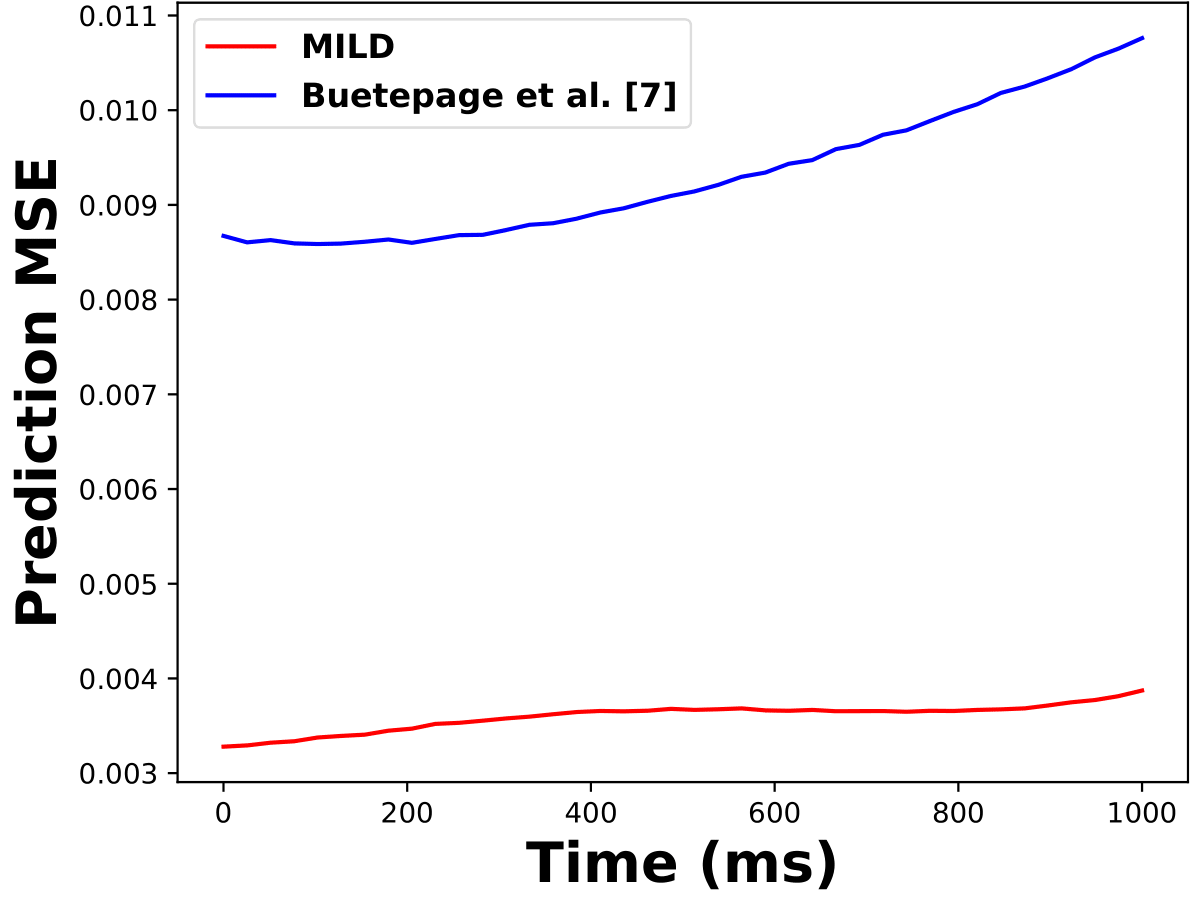}
        \caption{Handshaking}
        \label{fig:handshaking-pred}
    \end{subfigure}
    \begin{subfigure}[b]{0.24\textwidth}
        \centering
        \includegraphics[width=\linewidth]{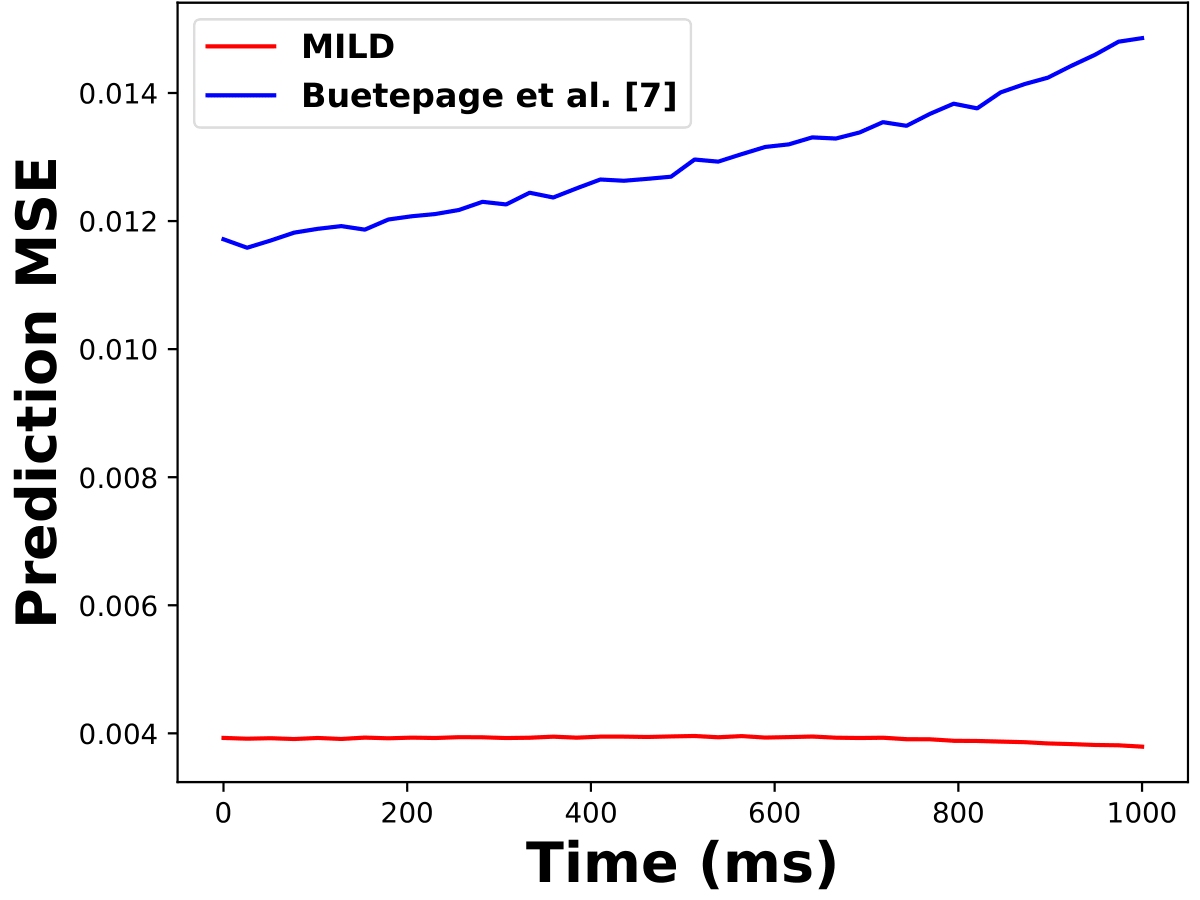}
        \caption{Rocket Fistbump}
        \label{fig:rocket-pred}
    \end{subfigure}
    \begin{subfigure}[b]{0.24\textwidth}
        \centering
        \includegraphics[width=\linewidth]{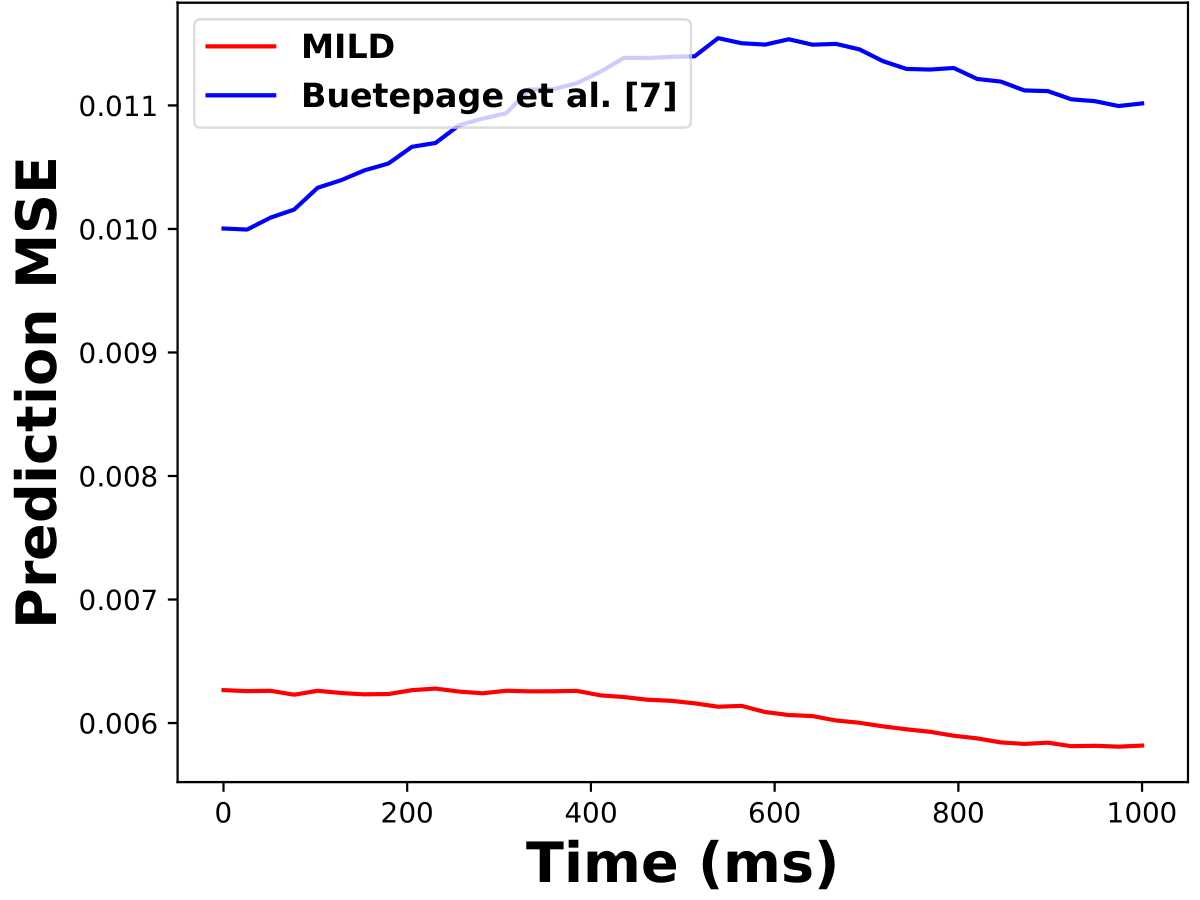}
        \caption{Parachute Fistbump}
        \label{fig:parachute-pred}
    \end{subfigure}
\caption{Prediction MSE of the second agent after observing the first agent averaged over 1 second (40 observations) of MILD compared with~\cite{butepage2020imitating}.
    (Lower is better, Red - MILD, Blue -~\cite{butepage2020imitating})}
        \label{fig:pred-mse}
        \vspace{-1em}
\end{figure*}


\begin{figure*}[h!]
\centering
    \begin{subfigure}[b]{0.24\textwidth}
        \centering
        \includegraphics[width=\linewidth]{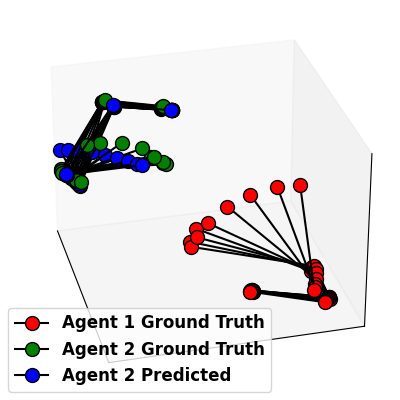}
        \caption{Hand Waving}
        \label{fig:handwave3d}
    \end{subfigure}
    \begin{subfigure}[b]{0.24\textwidth}
        \centering
        \includegraphics[width=\linewidth]{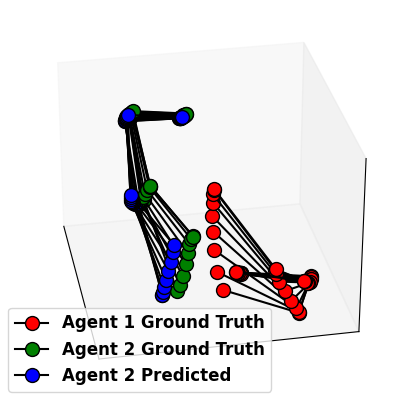}
        \caption{Handshaking}
        \label{fig:handshake3d}
    \end{subfigure}
    \begin{subfigure}[b]{0.24\textwidth}
        \centering
        \includegraphics[width=\linewidth]{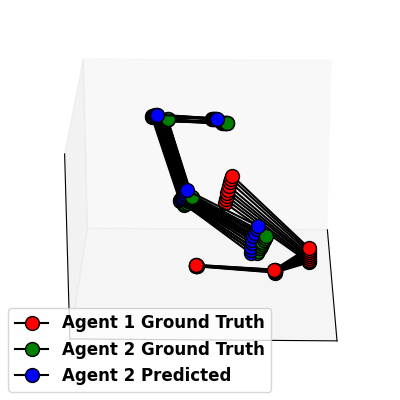}
        \caption{Rocket Fistbump}
        \label{fig:rocket3d}
    \end{subfigure}
    \begin{subfigure}[b]{0.24\textwidth}
        \centering
        \includegraphics[width=\linewidth]{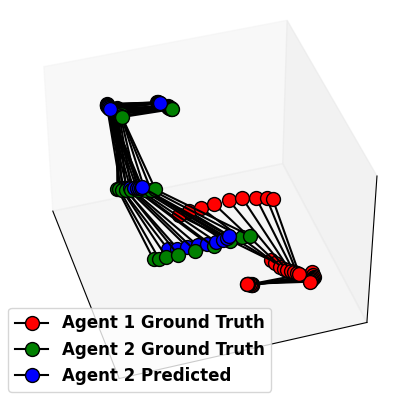}
        \caption{Parachute Fistbump}
        \label{fig:parachute3d}
    \end{subfigure}
    \caption{Sample Human-Human Interactions generated by MILD from the dataset in \cite{butepage2020imitating}. The 3D plots consist of 4 right arm joints of the interacting agents, along with the predictions of MILD of the second agent's joints after having observed the first agent. The points, both observed and predicted, correspond to a time window of 1 second (40 observations of which we visualize 8 for ease of vieweing). (Red - Ground Truth Trajectory of Agent 1, Green - Ground Truth Trajectory of Agent 2, Blue - Predicted Trajectory of Agent 2)}
        \label{fig:pred-butepage}
\end{figure*}

\begin{figure*}[h!]
    \centering
    \includegraphics[width=0.13\textwidth]{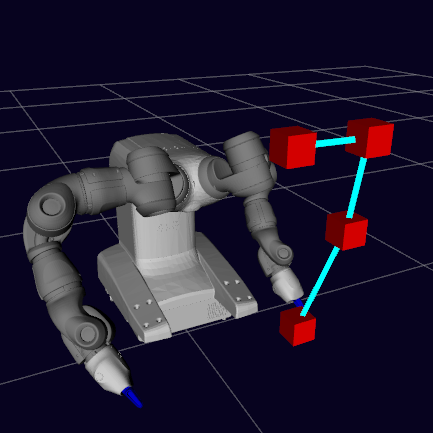} \includegraphics[width=0.13\textwidth]{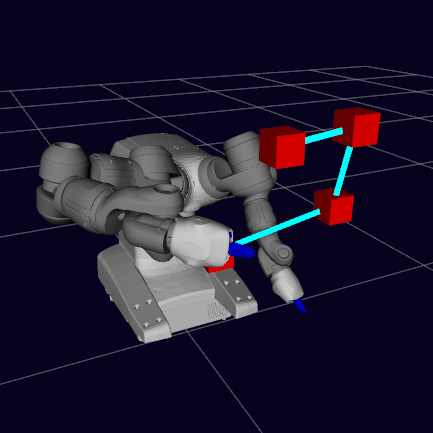} \includegraphics[width=0.13\textwidth]{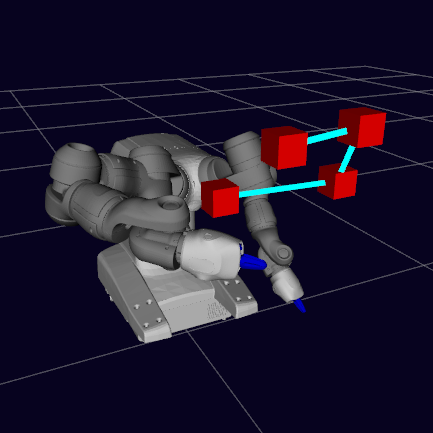} \includegraphics[width=0.13\textwidth]{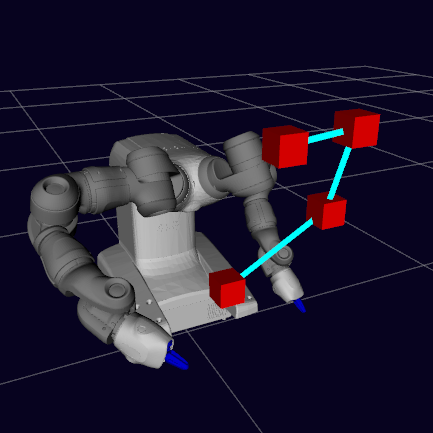}  \includegraphics[width=0.13\textwidth]{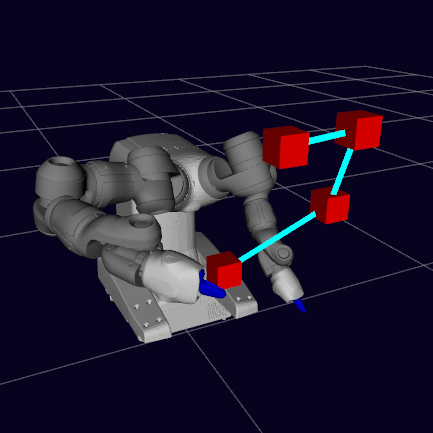} \includegraphics[width=0.13\textwidth]{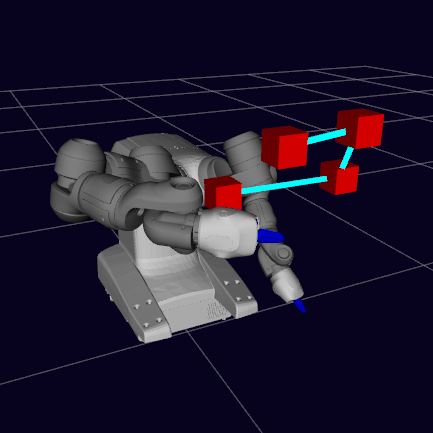} \includegraphics[width=0.13\textwidth]{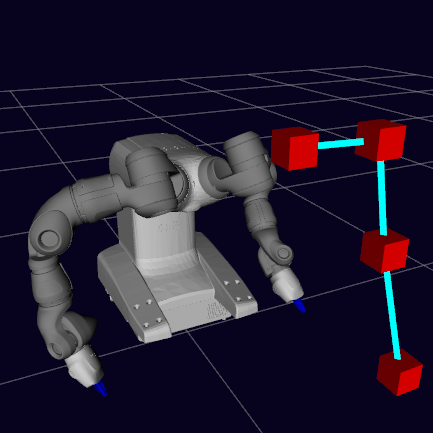}
    \caption{Example of a Handshaking HRI generated by MILD on the Yumi robot after training on \cite{butepage2020imitating}. The Red boxes are the observed right arm 3D coordinates of the human partner.}
    \label{fig:yumi-handshake}
\end{figure*}

Fig.~\ref{fig:latents} shows some examples of the latent space distributions learned by MILD on the HHI data of \cite{butepage2020imitating}. It can be seen that MILD is able to capture most of the different modes of the distribution, rather than forcing them to fit a standard normal distribution. Despite the relatively low spread of the data, the HSMM prior nicely encapsulates the shape of the trajectories, and learns the corresponding sequencing between the modes, thereby effectively capturing the dynamics of the trajectories, as in~\cite{nagano2019hvgh}. However, given the full structure of our covariance matrices, MILD is able to capture the interactivity between the agents as well.


\begin{table}[h!]
    \centering


\begin{tabular}{|cc|c|c|}
\hline
\multicolumn{2}{|c|}{Method}                                                                & MILD                           & \cite{butepage2020imitating} \\ \hline
\multicolumn{1}{|c|}{}& Wave          & \textbf{0.669 $\pm$ 0.910$^*$}              & 1.097 $\pm$ 1.719                             \\ \cline{2-4} 
\multicolumn{1}{|c|}{Prediction MSE}                                                 & Shake          & \textbf{0.360 $\pm$ 0.301$^*$} & 0.931 $\pm$ 0.728                            \\ \cline{2-4} 
\multicolumn{1}{|c|}{for Agent 2 (cm)}                                                 & Rocket    & \textbf{0.391 $\pm$ 0.586$^*$} & 1.295 $\pm$ 1.344                             \\ \cline{2-4} 
\multicolumn{1}{|c|}{}                                                 & Parachute & \textbf{0.610 $\pm$ 0.100$^*$} & 1.101 $\pm$ 0.668                             \\ \hline
\end{tabular}
\vspace{0.5em}
    \caption{Prediction MSE (in cm) of MILD compared with~\cite{butepage2020imitating} on the dataset in~\cite{butepage2020imitating} when predicting the trajectory of the second agent after observing the first agent. (Lower is better, * - $p<0.05$)}
    \label{tab:pred-buetepage}
    \vspace{-2em}
\end{table}

\begin{table}[h!]
\begin{tabular}{|cc|c|c|}
\hline
\multicolumn{2}{|c|}{Method}                                                                        & MILD                           & \cite{butepage2020imitating} \\ \hline
\multicolumn{1}{|c|}{} & Wave          & \textbf{0.084 $\pm$ 0.089$^*$} & 0.348 $\pm$ 0.363                             \\ \cline{2-4} 
\multicolumn{1}{|c|}{Prediction MSE for}                                                         & Shake          & \textbf{0.038 $\pm$ 0.035$^*$} & 0.371 $\pm$ 0.224                            \\ \cline{2-4} 
\multicolumn{1}{|c|}{the Yumi Robot}                                                         & Rocket    & \textbf{0.077 $\pm$ 0.056$^*$} & 0.362 $\pm$ 0.398                             \\ \cline{2-4} 
\multicolumn{1}{|c|}{(rad)}                                                         & Parachute & \textbf{0.067 $\pm$ 0.049$^*$} & 0.491 $\pm$ 0.224                             \\ \hline
\end{tabular}
\vspace{0.5em}
    \caption{Prediction MSE (in radians) for comparing our method, MILD, with~\cite{butepage2020imitating} on the HRI trajectories from~\cite{butepage2020imitating}. We predict the trajectory of the robot after observing the human partner. \\(Lower is better, * - $p<0.05$)}
    \label{tab:pred-buetepage-hr}
    \vspace{-2em}
\end{table}

\begin{table}[h!]
\centering
\begin{tabular}{|c|c|}
\hline
Action          & Prediction MSE (cm) \\ \hline
Clap Fist       & 0.315 $\pm$ 0.173   \\ \hline
Fist Bump       & 0.195 $\pm$ 0.137   \\ \hline
Handshake       & 0.173 $\pm$ 0.109   \\ \hline
High Five       & 0.293 $\pm$ 0.191   \\ \hline
Rocket Fistbump & 0.422 $\pm$ 0.495   \\ \hline
Hand Wave       & 1.325 $\pm$ 1.219   \\ \hline
\end{tabular}
\vspace{0.5em}
    \caption{Prediction MSE (in cm) of MILD on the NuiSI dataset when predicting the trajectory of the second agent after observing the first agent. (Lower is better)}
    \label{tab:pred-nuisi}
\vspace{-3em}
\end{table}

\begin{figure}[h!]
    \centering
    \includegraphics[width=0.8\linewidth]{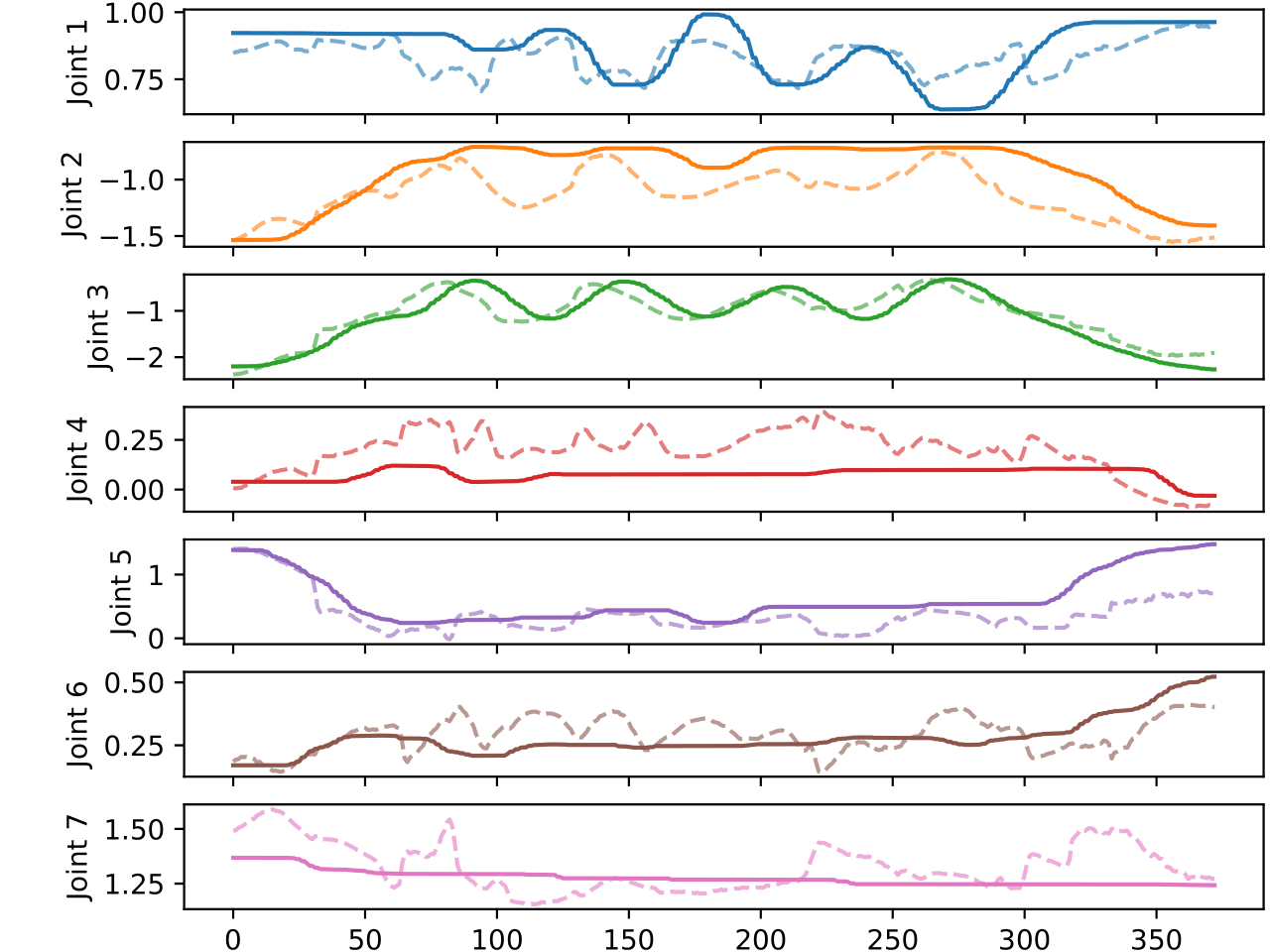}
    \caption{An example of a trajectory generated by MILD for a Handshaking HRI on the Yumi robot. The dotted lines show the predicted robot joint trajectories and the solid lines show the ground truth (X axis - Time steps, Y axis - Joint angles (radians)).}
    \label{fig:handshake-yumi-plot}
\end{figure}

\begin{figure*}[h!]
\centering
    \vspace{1em}
    \begin{subfigure}[b]{0.15\textwidth}
        \centering
        \includegraphics[width=\linewidth]{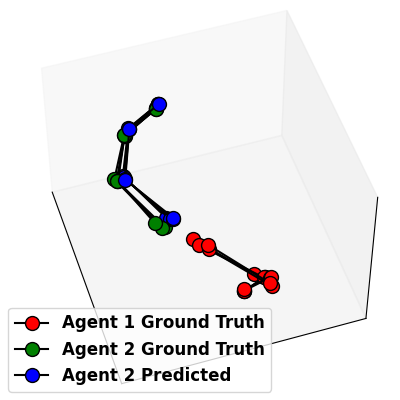}
        \caption{Clap Fist}
        \label{fig:nuisi-clapfist}
    \end{subfigure}
    \begin{subfigure}[b]{0.15\textwidth}
        \centering
        \includegraphics[width=\linewidth]{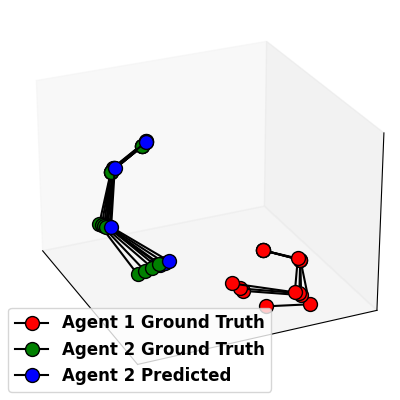}
        \caption{Fist Bump}
        \label{fig:nuisi-fistbump}
    \end{subfigure}
    \begin{subfigure}[b]{0.15\textwidth}
        \centering
        \includegraphics[width=\linewidth]{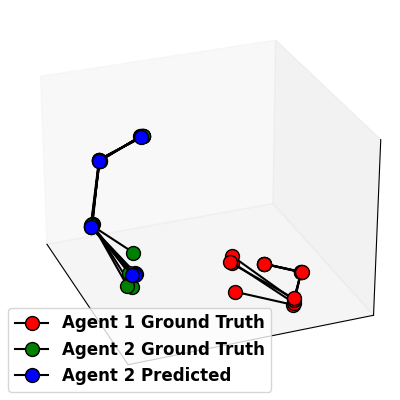}
        \caption{Handshake}
        \label{fig:nuisi-handshake}
    \end{subfigure}
    \begin{subfigure}[b]{0.15\textwidth}
        \centering
        \includegraphics[width=\linewidth]{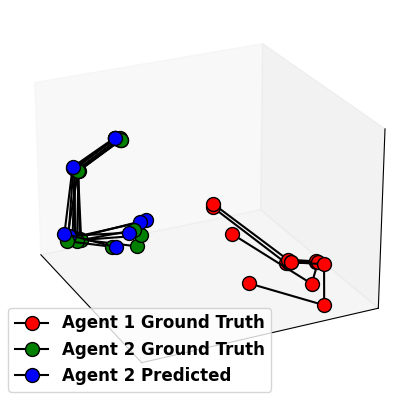}
        \caption{High Five}
        \label{fig:nuisi-highfive}
    \end{subfigure}
    \begin{subfigure}[b]{0.15\textwidth}
        \centering
        \includegraphics[width=\linewidth]{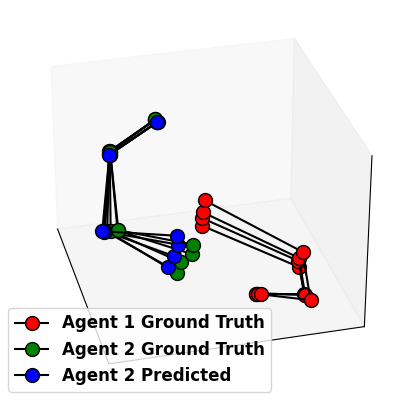}
        \caption{Rocket Fistbump}
        \label{fig:nuisi-rocket}
    \end{subfigure}
    \begin{subfigure}[b]{0.15\textwidth}
        \centering
        \includegraphics[width=\linewidth]{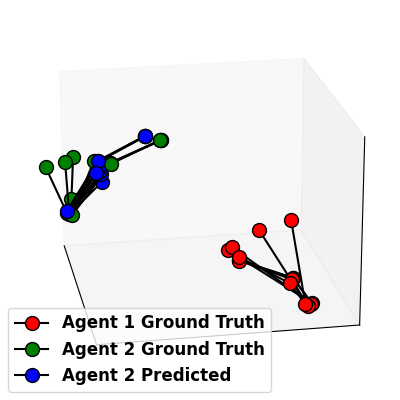}
        \caption{Hand Wave}
        \label{fig:nuisi-wave}
    \end{subfigure}
    \caption{Sample Human-Human Interactions generated by MILD from the NuiSI dataset. The 3D plots consist of 4 right arm joints of the interacting agents, along with the predictions of MILD of the second agent's joints after having observed the first agent. We show the observed and predicted values over 5 time steps. (Red - Ground Truth Trajectory of Agent 1, Green - Ground Truth Trajectory of Agent 2, Blue - Predicted Trajectory of Agent 2)}
        \label{fig:pred-nuisi}
\end{figure*}

\begin{figure*}[h!]
    \centering
    \includegraphics[width=0.16\textwidth]{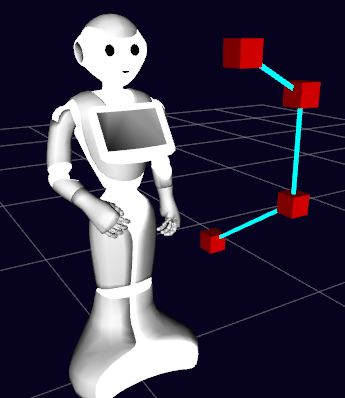} \includegraphics[width=0.16\textwidth]{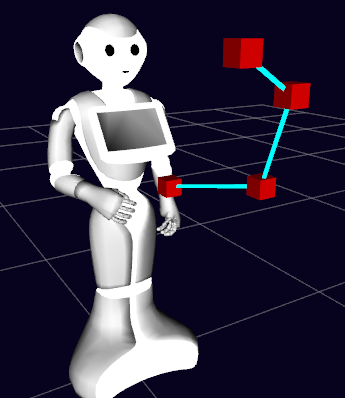} \includegraphics[width=0.16\textwidth]{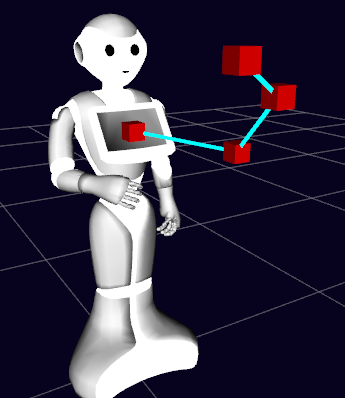} \includegraphics[width=0.16\textwidth]{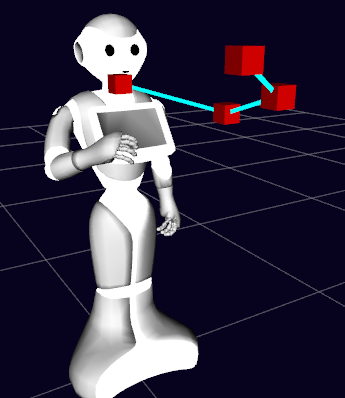} \includegraphics[width=0.16\textwidth]{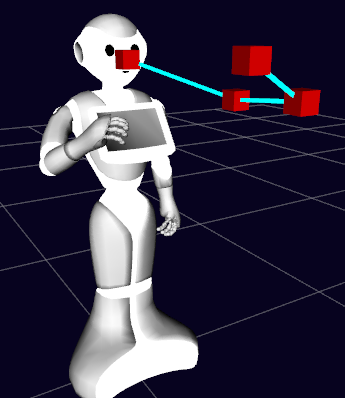} \includegraphics[width=0.16\textwidth]{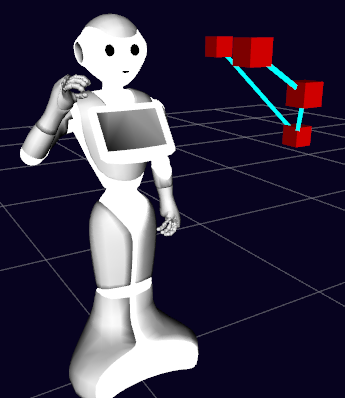} 
    \caption{Example of a Rocket Fistbump HRI generated by MILD on the Pepper robot after training on the NuiSI dataset. The Red boxes are the observed right arm 3D coordinates of the human partner.}
    \label{fig:pepper-rocket}
    \vspace{-2em}
\end{figure*}

\subsection{Conditioned Prediction Results}
\label{ssec:motion_pred}

We test the conditioning ability of MILD compared to~\cite{butepage2020imitating}\footnote{Results reported using our own implementation of~\cite{butepage2020imitating}.} to evaluate the accuracy of the generated motions of the controlled agent after observing the human interaction-partner. We evaluate the approaches over the four interactions of the dataset in \cite{butepage2020imitating}. We calculate the mean squared error (MSE) between the predicted motions and the ground truth skeleton/robot joints at each time step after observing the human's trajectory. The results of these interactions can be seen in Fig.~\ref{fig:pred-mse} and Table~\ref{tab:pred-buetepage}. We observe that MILD consistently performs better than~\cite{butepage2020imitating} and accurately predicts the motions of the interacting agent over all the different interactions. One specific reason behind this could be that in~\cite{butepage2020imitating}, they learn a common spatial latent space followed by a dynamics latent space, regularizing the spatial embeddings of each agent to match temporally. During testing, this could create errors when generating from the temporal latent space, since there is no way to distinguish whether a given sample corresponds to the first agent or the second agent. This shows the power of the explicit conditioning ability of HSMMs in predicting interactive behaviors since the agents can be jointly modeled, as compared to learning just a single representation over the two agents. Some qualitative results of the predicted skeletons from the dataset in~\cite{butepage2020imitating} is can be seen in Fig.~\ref{fig:pred-butepage}, where the trajectories of both the interacting agents over a 1 second period (8 out of 40 time steps) and the predictions of MILD for the second agent are visualized.

We additionally show the results of training on the HRI trajectories from the dataset in~\cite{butepage2020imitating}. In this case, the accuracy obtained using MILD (Table.~\ref{tab:pred-buetepage-hr}) outperforms ~\cite{butepage2020imitating} over all the interactions, showing that our method is able to effectively model not just HHI tasks but also HRI tasks. However, the general accuracy is not as good as the HHI scenario, as can be seen in Fig. \ref{fig:handshake-yumi-plot}. This can be attributed to the relatively small size of the dataset (only 32 HRI trajectories are present in the dataset, compared to 149 HHI trajectories). In such cases, learning HRI from HHI can yield a more viable solution. Despite this, we are able to nicely capture the temporal dependencies between the human and the robot, which can be seen in the interactivity of an example Human-Robot handshaking exchange in Fig. \ref{fig:yumi-handshake}. Here, we show how an HRI between a simulated Yumi robot and the observed (simulated) human, whose right arm joints are shown in red. It can be seen that the motions of the robot are in sync (within the 1 second delay) with that of the human during the handshake, which shows that MILD is able to model the interactivity in such tasks.

On the NuiSI dataset, MILD is able to achieve good accuracy (within cm range as shown in Table~\ref{tab:pred-nuisi}) in predicting the skeleton trajectories of the second (controlled) agent after observing the first agent. Examples of these trajectories can be seen in Fig.~\ref{fig:pred-nuisi} where one can see the accuracy of our predicted skeletons (in blue). This shows that MILD performs well not just on idealistic motion capture data, but also on real-world interaction scenarios involving relatively noisy inputs. We additionally show some quantitative results of MILD on NuiSI by mapping the learned trajectories to a Pepper Robot. We do so by making use of the similarities in the DoFs between a human arm and Pepper, using which the joint angles are calculated from the generated skeletons, which are then executed on the robot~\cite{fritsche2015first,prasad2021learning}. An example of a Rocket Fistbump HRI with a simulated Pepper can be seen in Fig.~\ref{fig:pepper-rocket}.  Please refer to our supplementary video and site for the real-robot demonstration.

\section{Conclusion and Future Work}
\label{sec:conclusion}
In this paper, we proposed an extension to the general idea of using Learning for Demonstration (LfD) techniques for learning latent space dynamics from human observations. We proposed Multimodal Interactive Latent Dynamics (MILD), a method that combines Hidden Semi-Markov Models (HSMMs) as prior in  Variational Autoencoders (VAEs) for learning Human-Robot Interaction. The predictive ability of HSMMs, allows us to effectively model a joint distribution over the demonstrated trajectories from Human-Human interactions. We can generate robot control trajectories during test time by conditioning HSMMs on the human-partner actions in a timely and reactive manner. Our approach performs better than the current state-of-the-art for learning latent interaction dynamics in HRI, which we show through extensive experiments on a diverse range of datasets.

Currently, we do not incorporate any action recognition for selecting the corresponding latent HSMM that needs to be executed. For our future work, we would first focus on either incorporating an action recognition framework for this or exploring the modular ability of HSMMs by increasing the number of Gaussian Components to learn different interactions using a single HSMM, which can then be used for looking into trajectory segmentation as in~\cite{krishnan2017transition,lioutikov2015probabilistic}. 
Moreover, we plan to incorporate Inverse Kinematics to better improve the physical contact during the interaction, rather than looking just at the joint space trajectories. 

\section*{Acknowledgements}
The authors would like to thank Mark Baierl, Michel Kohl and Martina Gassen whose work helped in developing of parts of the approach. 

\bibliographystyle{IEEEtrans}
\bibliography{IEEEexample}  

\end{document}